\documentclass[review]{elsarticle}
\let\today\relax
\makeatletter
\def\ps@pprintTitle{%
    \let\@oddhead\@empty
    \let\@evenhead\@empty}
 
\makeatother

\def \pfName {Nemesyst}
\def \pTitle {\pfName{}: A Hybrid Parallelism Deep~Learning-Based Framework Applied for Internet of Things Enabled Food Retailing Refrigeration Systems}
\usepackage{datetime2} % sets datetime style to comp sci
\usepackage{graphicx}
\usepackage{rotating} % for 'sidewaystable' env.
\graphicspath{ {./content/images/} }
\usepackage[table]{xcolor}% http://ctan.org/pkg/xcolor
\usepackage{booktabs}
\usepackage{lscape}
\usepackage{hyperref}
\usepackage{siunitx}
\usepackage{adjustbox}
\usepackage{geometry}
\usepackage{chngcntr}
\usepackage{setspace}
\usepackage{textcomp}
\begin{document}
 \title{\pTitle}

% The paper headers % what is this?
\markboth{Journal of \LaTeX\, \today}%
{Shell \MakeLowercase{\textit{et al.}}: Bare Demo of IEEEtran.cls for Computer Society Journals}
\author[add1]{George Onoufriou}
\author[add2]{Ronald Bickerton}
\author[add3]{Simon Pearson}
\author[add1]{Georgios Leontidis\corref{cor1}}
\ead{gleontidis@lincoln.ac.uk}

\begin{abstract}

Deep Learning has attracted considerable attention across multiple application domains, including computer vision, signal processing and natural language processing. Although quite a few single node deep learning frameworks exist, such as tensorflow, pytorch and keras, we still lack a complete processing structure that can accommodate large scale data processing, version control, and deployment, all while staying agnostic of any specific single node framework. To bridge this gap, this paper proposes a new, higher level framework, i.e. Nemesyst, which uses databases along with model sequentialisation to allow processes to be fed unique and transformed data at the point of need. This facilitates near real-time application and makes models available for further training or use at any node that has access to the database simultaneously. Nemesyst is well suited as an application framework for internet of things aggregated control systems, deploying deep learning techniques to optimise individual machines in massive networks. To demonstrate this framework, we adopted a case study in a novel domain; deploying deep learning to optimise the high speed control of electrical power consumed by a  massive internet of things network of retail refrigeration systems in proportion to load available on the UK National Grid (a demand side response). The case study demonstrated for the first time in such a setting how deep learning models, such as Recurrent Neural Networks (vanilla and Long-Short-Term Memory) and Generative Adversarial Networks paired with Nemesyst, achieve compelling performance,  whilst still being malleable to future adjustments as both the data and requirements inevitably change over time.

\end{abstract}


\cortext[cor1]{Corresponding author. Tel.:+44(0)7951053731}

\address[add1]{School of Computer Science, University of Lincoln, Brayford Pool Campus, LN67TS, Lincoln, UK}
\address[add2]{School of Engineering, University of Lincoln, Brayford Pool Campus, LN67TS, Lincoln, UK}
\address[add3]{Lincoln Institute for Agri-Food Technology, University of Lincoln, Brayford Pool Campus, LN67TS, Lincoln, UK}

\maketitle

\section{Introduction}\label{sec:introduction}

Intelligence derived from data has the potential to transform industrial productivity. Its application requires computational solutions that integrate database technologies, infrastructure and frameworks along with bespoke machine learning techniques to handle, process, analyse, and recommend. Recent advances in deep learning are accelerating application.  

Deep learning is a sub-field of machine learning that seeks to learn data representations, usually via a variant of artificial neural networks. Deep learning is concerned with two distinct tasks:
\begin{itemize}
    \item Training; where the deep neural network (DNN) learns the data representation through iterative methods, such as stochastic gradient descent. This is a very time consuming operation that usually occurs at most on one node, or a few CPU /GPU nodes.
    \item Inference; where the learned DNN can be used in one form or another to make predictions, detection, localisation, etc. This is usually faster than training as it does not require as many operations, but still requires a good portion of the same resources and time. \cite{panda2019high}
\end{itemize}

Both training and inference can be improved by the use of 'better' hardware, including faster and/or higher volume processing via distributed/parallel systems, where parallel in this instance means any single system that according to Flyns taxonomy is using either:
\begin{itemize}
    \item single instruction, multiple data (SIMD)
    \item multiple instruction, single data (MISD)
    \item multiple instruction, multiple data (MIMD)
\end{itemize}
This is separate from a distributed system, where each computing node is a different networked computer, as opposed to a single computer with multiple CPUs/GPUs for computation in parallelisation. Currently the majority of deep learning frameworks have matured in parallelisation, but they lack or are still in their infancy with regard to distribution. Novel solutions such as TensorFlow-distribute, or IBMs distributed deep learning are now emerging to resolve this issue \cite{panda2019high}. The ability to share data between distributed nodes using deep learning frameworks is a critical bottleneck, despite the prevalence of databases providing very mature and advanced functionality, such as MongoDB's aggregate pipeline. In addition, the unification of deep learning interfaces to databases will aid reproducibility, as it will standardise the message protocol interface (MPI) and describe what data in what manner is used for training and inference; in MongoDBs case, by using aggregate pipeline json files. The integration of deep learning into database models is crucial for any (near) real-time application, such as Netflix, Amazon, Google, as 'rolling' a bespoke data management system or MPI is costly and can result in inferior performance and capabilities.\par 

In this paper, we propose a novel end-to-end distributed framework that can handle any type of data, scales and distributes models across various nodes locally or on the cloud. We demonstrate the approach by developing a deep learning infrastructure to link the power control of a massive internet of things (IoT) enabled network of food retail refrigeration systems in proportion to the load available on the UK National Grid (i.e. an optimised Demand Side Response, DSR). The implementation utilises data from a large retail supermarket along with a purpose-built demonstration/experimental store (the  'barn') in our Riseholme Campus at the University of Lincoln.
 \section{Background}

\subsection{Deep Learning Frameworks}
Deep Learning, and its ability to learn complex abstract representations from high dimensional data, is transforming the application of artificial intelligence.  Deep learning can learn a new task from scratch, or adapt to perform tasks in various domains by transferring knowledge from one domain to another. Areas where deep learning, including Capsule Networks \cite{sabour2017dynamic, ribeiro2019capsule}, has produced state-of-the-art results -- and why we consider it here -- are in particular; computer vision, and medical imaging \cite{esteva2019guide, kollias2017adaptation, ribeiro2018end, nie2018deep, DeSousaRibeiro2019}; signal, and natural language processing \cite{ye2018power, zhang2018survey, ribeiro2018towards, young2018recent}; agriculture \cite {kamilaris2018deep, barbedo2019plant, alhnaity2019using, ringland2019characterization}; and industry \cite{zhang2018efficient, li2018deep, zhang2018tensor}. 

There are many popular single node deep learning frameworks currently available \cite{tensorflow2015-whitepaper, jia2014caffe, paszke2017automatic, chollet2018keras}. However,  there are fewer parallelisable deep learning frameworks \cite{tensorflow2015-whitepaper, seide2016cntk}. These can broadly be categorised by their parallelism strategies and capabilities, namely: \cite{panda2019high}
\begin{itemize}
    \item Model parallelism (MP) (TensorFlow, Microsoft CNTK), where a single DNN model is trained using a group of hardware instances, and a single data set.
    \item Data parallelism (DP), where each hardware instance trains across different data (usually separate models).
    \item Hybrid parallelism (HP) (\pfName), where a group of hardware trains a single model, but multiple groups can be trained simultaneously with independent data sets simultaneously.
    \item Automatic selection (AP), where different parts of the training / inference process are tiled, with different forms of parallelism between tiles.
\end{itemize}

Of note, TensorFlow v1 uses a model parallelism strategy as it trains a single model over multiple GPUs on the same data. TensorFlow has recently added functionality called TensorFlow distribute which adds the ability to train on multiple machines connected by a network \cite{tensorflow.distribute} however this remains model parallelism as it requires that the entire set of machines broadcast replicas of the data, such that it does not use different data in a single process itself and instead requires that it be run over multiple processes externally or other such methods like parameter servers to simulate data parallelism. \cite{tensorflow.distribute} Their current method for multi machine data broadcasting can be seen in what TensorFlow distribute uses, what they call collective\_ops using methods such as all reduce as can be seen in their source code \cite{tensorflow.collectiveops} which has limited functionality compared to a more specialised tool such as MongoDB.

In contrast \pfName\ is a hybrid parallelism framework as it can simultaneously wrangle, learn, and infer multiple models, with multiple and distinct data sets, across a distributed/ wide system. In addition, \pfName\ can utilise existing single node frameworks, such as Keras \cite{chollet2018keras}, TensorFlow, PyTorch, since training scripts are separate, distinct, and configurable. The  \pfName\ codebase also includes examples and inbuilt functionality for multiple deep learning methods, such as RNN, LSTM, and GANs. \pfName's development was inspired, tested, and accommodated by the type of data and problems we encountered through the handling of real data sets, including and not limited to: too-large-to-fit data, requiring out-of-core processing; Distributed points of need; requiring some method of near-real-time dissemination; Changing/ dynamic data, requiring dynamic data handling for ever changing scenarios, and scenario subsets such as some stores using $CO_{2}$ refrigerators vs water vs some other coolant. All of which are not unusual circumstances for many large industries to face.

\subsection{Case Study - Demand Side Response Demonstration}
\begin{figure}[t!]
\centering
\hspace*{-.4cm} 
\includegraphics[width=0.8\columnwidth]{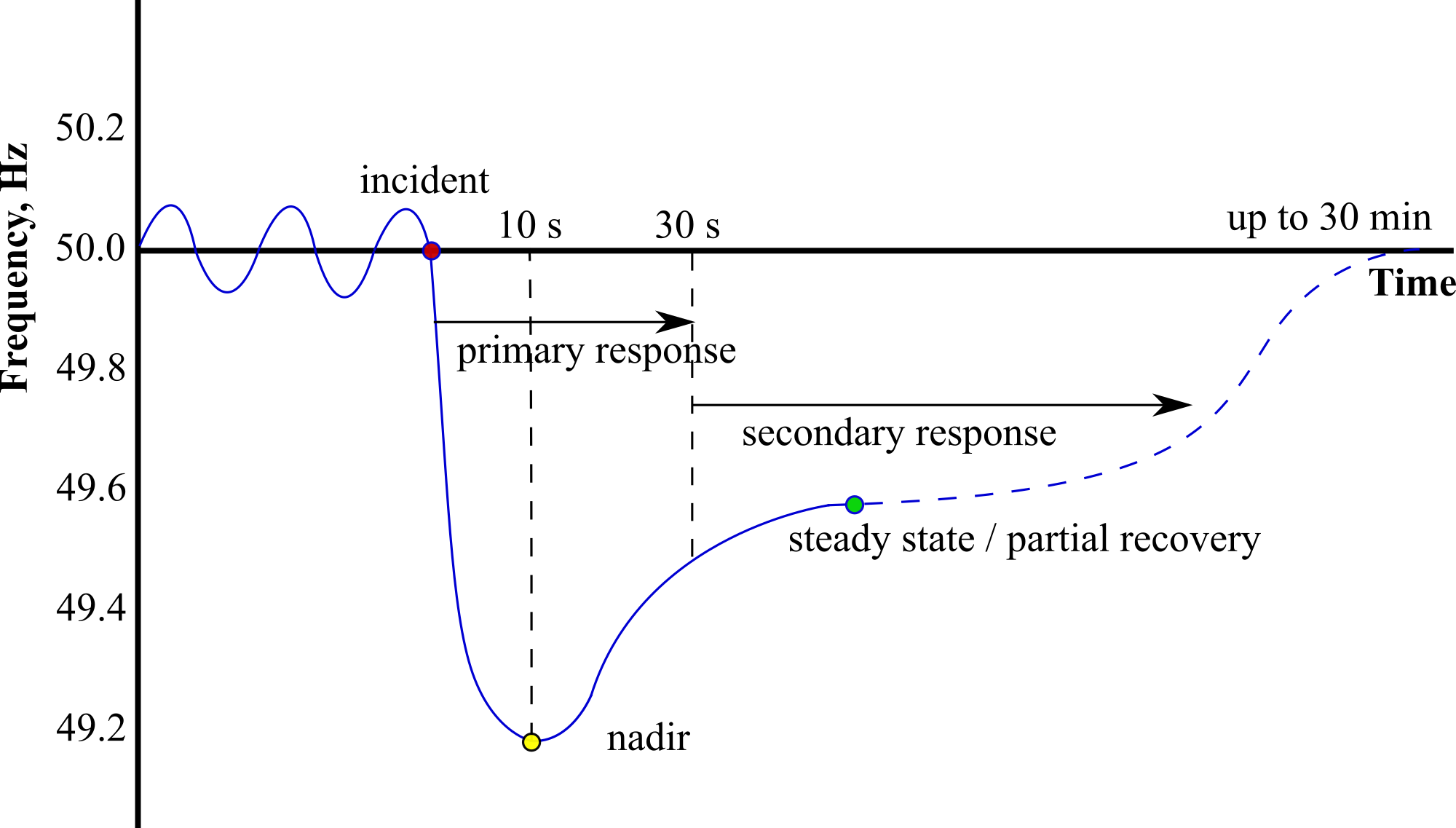}
\caption{(colour online) Firm Frequency Response overview (National Grid).}
\label{fig:ffr_overview}
\end{figure}

Here we demonstrate \pfName\ by optimising the energy use and system operation of a massive IoT network of retail refrigeration systems. High demands for energy, along with future commitments towards low carbon economies that include the  generation of less predictable electrical loads (e.g. wind / solar renewables) is placing National Grid infrastructure under increased strain. To stabilise the Grid,  focus is now placed on developing new approaches to manage load via the application of Demand Side Response (DSR) measures. During DSR events individual electrical demand across massive aggregated networks of machines is adjusted in response to available grid supply. Measures span financial incentives for commercial/industrial users to reduce demand during peak times when there is low available power and vice versa \cite{bradley2016financial}.  Many DSR incentives require the control of large loads and the aggregation of multiple  industrial processes and machines \cite{granell2016power, grunewald2013demand}. Standard static DSR mechanisms includes Firm Frequency Responses (FFR), where load is shed when the grid frequency drops to a predefined threshold. Figure \ref{fig:ffr_overview} shows the two transitions typically encountered in a static FFR event. In the first transition (also known as primary FFR), industry has to shed load quickly (2  to 10s) and sustain this for 30 seconds; for the second transition the load can be held off for 30 minutes (also known as secondary FFR). Primary FFR's aim is  to arrest frequency deviations from the nominal value, whereas secondary FFR is performed to contain and partially recover the frequency \cite{teng2015benefits}. \par 

        The application of DSR to massive networks of machines requires exceptional levels of coordinated control across individual devices. Food refrigeration systems can be deployed within DSR events since the thermal inertia / mass of food acts as a store of cold energy (effectively a battery). However, the thermal inertia in a retail refrigeration case will change as the food is actively shopped by consumers and then refilled, in addition ambient temperature can change, networks of stores can have multiple refrigeration systems and different foods will need bespoke control.  This complexity provides an excellent opportunity for \pfName\ since deep learning is required to model thousands of assets simultaneously, with a need to continuously retrain multiple thousands of models / machine dynamics. The deployed control model must intelligently select which refrigerators to shed electrical load / turn off across a massive pool; one UK retailer alone operates $>$100,000 devices. This requires an algorithm that can predict the thermal inertia in individual cases and therefore how long they can be shut down (typically up to a maximum of up to 30 minutes) before the food temperature breaches a legally defined set point. Deep learning continuously re learns the thermal inertia of each device.  To deliver DSR, a seperate candidacy algorithm selects which refrigerators can be shut down across an aggregated pool and for how long. For compliant DSR, a fixed electrical load, typically greater then 10MW, must be shed across the entire pool of machines for up to 30 minutes.
        
        More generally, the application of deep learning for refrigeration system control is limited, for example  Hoang \cite{DBLP:journals/corr/abs-1812-00679} purports gains of 7\% in daily chiller energy efficiency by using deep learning opposed to rote physical control models. There have also been some studies focussing on load forecasting for home appliances and energy disaggregation including refrigerators \cite{kong2018short, guo2019drivers, hossain2019application}. Although quite a few studies exist in related areas of research, such as demand response algorithms for smart-grid ready residential buildings in \cite{pallonetto2019demand} or demand response strategies in autonomous micro-grids in \cite{panagiotidis2019r}, there has been no studies specifically deploying deep learning for DSR control in food retailing refrigeration systems.

\begin{figure}[!hbt]
    \centering
    \includegraphics[width=\columnwidth]{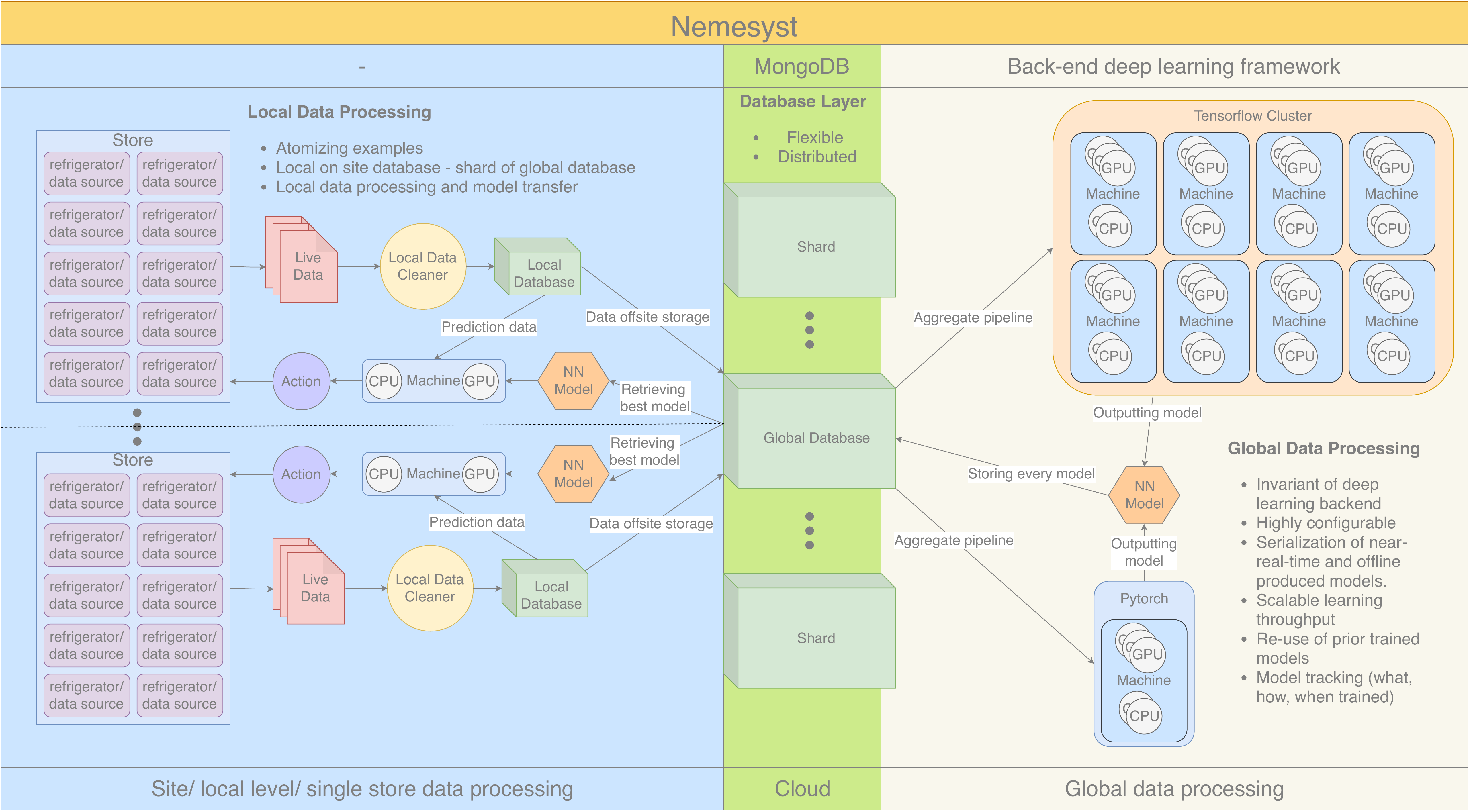} % the updated text + size version
	\caption{(colour online) The literal representation of \pfName\ when applied to distributed refrigeration, in contrast to \ref{fig:framework} which instead shows an abstraction of the Nemesyst framework. \cite{nemesyst.readthedocs}}
    \label{fig:framework_literal}
\end{figure}
       
 \section{Nemesyst Framework}
\label{sec:framework}

The goal of this framework (figure 2) is to facilitate the use of deep learning on large distributed data sets in a  distributable, accurate, and timely manner. As such, each stage must consider the following criteria:

\begin{itemize}
    \item Handle data out-of-core as the incoming data may be too large to store in memory at any single instance
    \item Retrieve necessary data from locations that may differ to where the local framework process exists
    \item Disseminate models to the agents that require them for inference.
\end{itemize}
\begin{figure}[!hbt]
    \centering
    \includegraphics[width=\columnwidth]{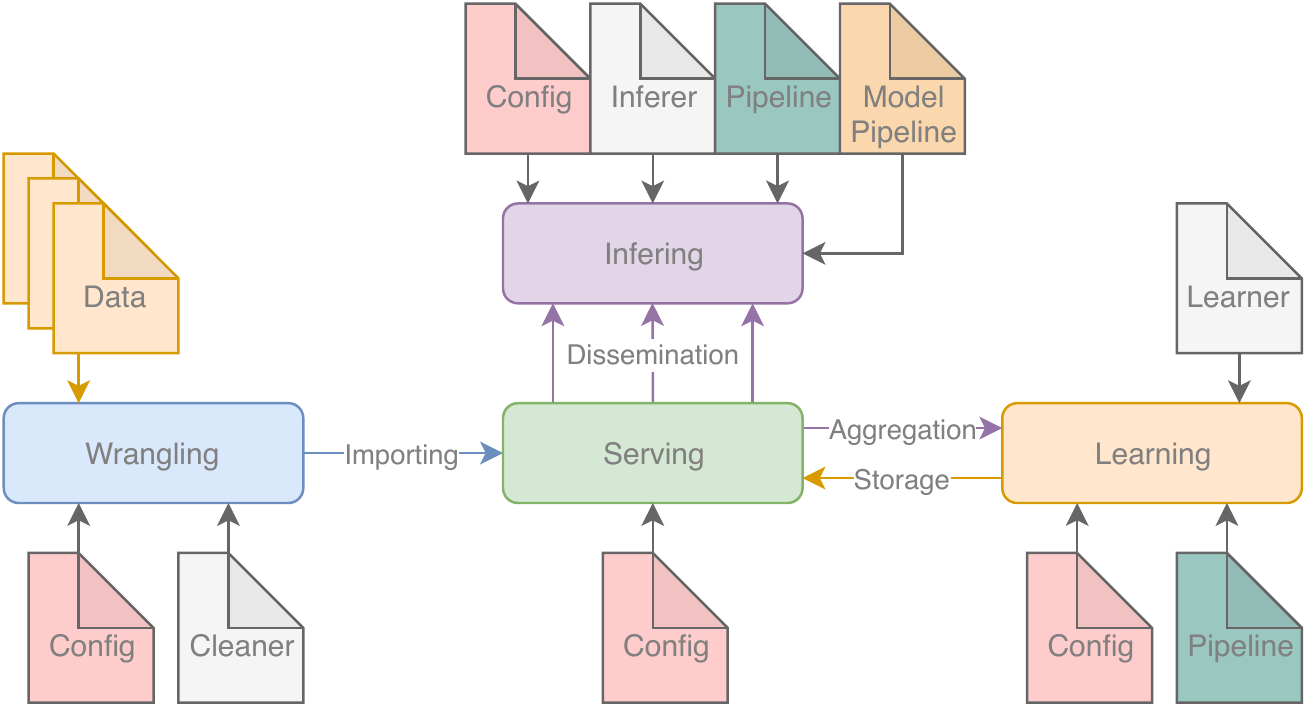}
	\caption{(colour online) The proposed high level framework architecture, in contrast to \ref{fig:framework_literal} which instead shows the literal representation of this abstraction. \cite{nemesyst.readthedocs}}
	\label{fig:framework}
\end{figure}
To achieve our goals and also be more agnostic and generalisable with respect to deep learning problems and techniques, we propose a new framework primarily consisting of four abstract core components. While each component is not new in of themselves combining them together as a formal framework to enable distributed deep learning in a hybrid parallel manner, agnostic of deep learning back-end is (figure \ref{fig:framework}).

\subsection{Wrangling}
     Wrangling; where the original data from a source is reformed into individual atomised/single usable examples, which will later be imported into a database such as MongoDb. This stage seeks to filter out the unusable data and to clean the remaining data from artifacts and inconsistencies. 
    
    As long as the data comprise single usable examples, e.g. in many-one sequence prediction multiple rows with a target column of a consistent value, in many sequence generation multiple rows with multiple values in the target column, etc., the data can be selectively chosen through MongoDB's aggregation pipelines. This approach allows for maximal flexibility and also generalises well, since a single example can be abstracted by the implementer via custom config and cleaning scripts. Since databases are not tailored specifically to deep learning and they are widely used, this forces us towards existing standards the benefit of which is that we can then utilise already existing frameworks for databases like Amazons DynamoDB for MongoDB and then further exploit the capabilities of these frameworks to manage our databases for use with in-place encryption and transport layer security (TLS).
    
    Data wrangling/ handling/ pre-processing as a concept is not new as of itself, and is prevalent throughout the deep learning, machine learning, and statistics communities.
    
\subsection{Serving}
 Serving; where clean and atomised data exists and can be stored such that it is readily available for the learners to access at the point of need; and as a consequence of being stored in a database can now be properly indexed. Subsequently indexing the data using a database means we are now free to utilise the advanced functionality of these databases including the aggregation pipelines, sharding (critical to distribution), searching, sorting, filtering, finding/ querying, and replica sets (critical for up-time and availability). 
    
    When a database is used in this manner, i.e separate the data cleaning and data storage from any specific learning script, the user creates a consistent interface for the learning scrips.  Unlike most deep learning scrips, our system is capable of branching many different and new algorithms as the server and learner are separate, this becomes the message passing interface (MPI). This stage also serves any trained models as learners are expected to return serialised binary objects of the learned weights and architecture.
    
Serving is not a new concept in many industries, but it is rarely a consideration in the deep learning community, left more as an afterthought focusing more on absolute model performance over usability, and practicality.
    
\subsection{Learning}
Learning; where the aggregated data is fetched in rounds to train a learner in batches, at which point the learner returns the training loss, validation loss, neural network architecture (sequential and binarised) to the serving database for dissemination for further learning, and use. At this stage it is possible and advisable to take advantage of other frameworks such as tensorflow, and possibly their distribute simultaneous training functionality on single machines.
The learning stage should be highly and completely configurable as it should not matter how a model is trained as long as it is returning all components desired in future, such as the trained model, and/ or training parameters.

\subsection{Inferring}
Inferring; where a trained serialised model stored in the database, and new data are aggregated and used to infer/ predict the target feature; What is represented by the distribution of the data. Usually toward some desired action or recommendation.

\subsection{Architecture Consideration Towards Goals}

We have implemented a form of this framework, dubbed: Nemesyst, \cite{nemesyst.repo} which incorporates deep learning techniques, such as RNNs, LSTMs, and GANs, and demonstrated its applicability and impact on a large scale industrial application of high impact. We have also released the code for this framework implementation under a permissive licence and will continue to add to it at: \\ https://github.com/DreamingRaven/Nemesyst.
 \section{Methods and Tools}
For the sake of both completeness and to maintain compactness this shall only be a brief look at the most important tools used.

\begin{itemize}
    \item Keras and Tensorflow;\cite{chollet2018keras} De facto standard for production level deep learning, utilising graphics processing unit (GPU) parallelisation to speed up computation. Both are free/ open-source removing the need for expensive corporate licensing
    \item MongoDb; perhaps the most popular noSQL database type, being document oriented, scalable, eventually consistent, and most importantly free/ open-source, and fast \cite{dede2013performance} 
\end{itemize}

\subsection{Parallelism/ Multiprocessing}

\begin{figure}[!hbt]
    \centering
    \includegraphics[width=\columnwidth]{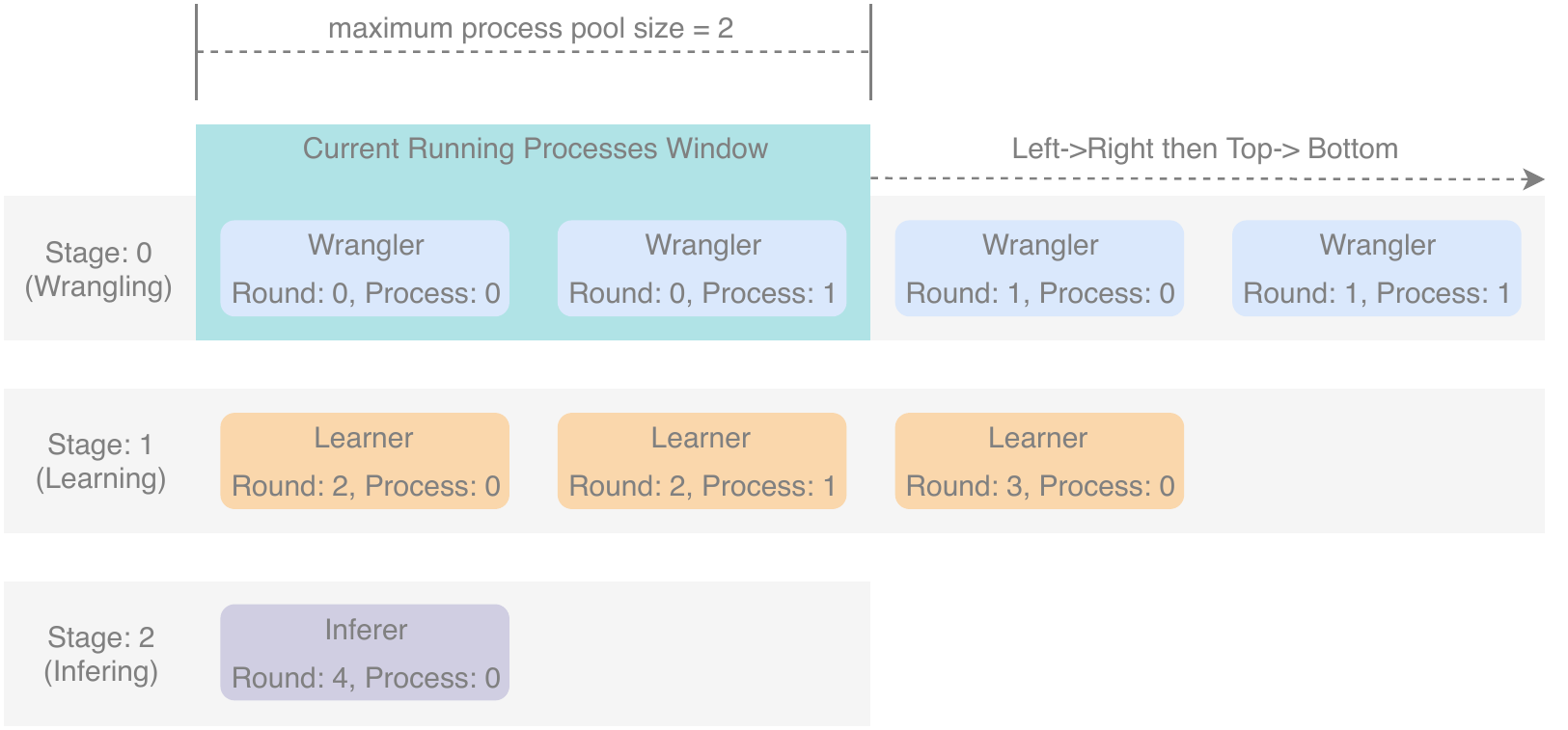}
	\caption{(colour online) Shows the multi-process strategy of \pfName\ on any single machine after it has been triggered. \cite{nemesyst.readthedocs}}
    \label{fig:framework_parallelism}
\end{figure}

 \pfName\ uses pythons multiprocessing module to implement parallelism through process pools which can be seen in figure \ref{fig:framework_parallelism}. Python enforces what is called the global interpreter lock (GIL), which means it is impossible to multi-thread python bytecodes as the GIL protects these bytecodes from access from multiple threads. In python the solution to this problem is to duplicate the to-be-multithreaded code and objects such that each process has its own independent copy which cannot affect the original or each other. This clearly involves overhead as objects must be duplicated but it also means there is little room for race conditions as they also operate completely independently. \par
 
 Each script/ algorithm provided to \pfName\ on a per-stage basis is run in parallel up to the maximum number defined by the user for the process pool size/ width. If more algorithms are provided than there are free processes then the process-window will slide across upon completion of the current running algorithms. The process window will not take algorithms from the next stages to fill any free processes as they could potentially depend on the outcome of the user defined algorithm in the current or previous stages. Only once all algorithms of a given stage have completed will the process pool move to the next stage if the user has chosen for \pfName\ to do so. 
 
 To facilitate parallel operations, each individual script is passed arguments to help it identify itself similar to how OpenCL's processing elements have access to their processing element number, along with other helpful statistics such as how many other processing elements of that round there are, how wide the window is, along with certain arguments being providable to each processing element individually.

\subsection{MongoDB}

In this framework we propose and use MongoDB as the data storage method. This comes with a plethora of consequences:
\begin{itemize}
    \item Distribution; utilising a client-server model through MongoDB affords the possibility to process individual models for individual tasks from a central database of models and/or data. In this case training a broader model for all refrigerators, which is then re-trained/ fine tuned individually in a distributed manner for individual machines
    \item Handling of large/ sharded data; Using a database allows much more complex data storage and representation to be handled, sorted, and maintained
    \item Querying; MongoDB is likely more effective at querying than bespoke system, as it is established, widely used and prevalent, meaning more support is available from the community itself
    \item Integration; due to MongoDB being a popular database, it is likely that this framework's MongoDB handling will be able to integrate to existing systems
    \item Complexity; handling any external and potentially changing framework can mean that the changes could cause unhandled exceptions, and temporary down time, until the differences can be adapted to
    \item Security; MongoDB is more mature and has many more available and used security implementations, for a variety of different scenarios, as will likely be the case upon application of this framework. This adaptability to the security need is an important part of the application of deep learning that is frequently ignored. Examples of such are in-place encryption, and transport layer security integration.
    \item Adaptability; the MongoDB aggregate pipeline and its generic nature has been included in this framework. This improve the application breadth. We note though that there must always be a target feature/ ground truth to be predicted along with the data to use for prediction.
\end{itemize}

\begin{figure}[hbt!]
    \centering
    \includegraphics[width=0.7\columnwidth]{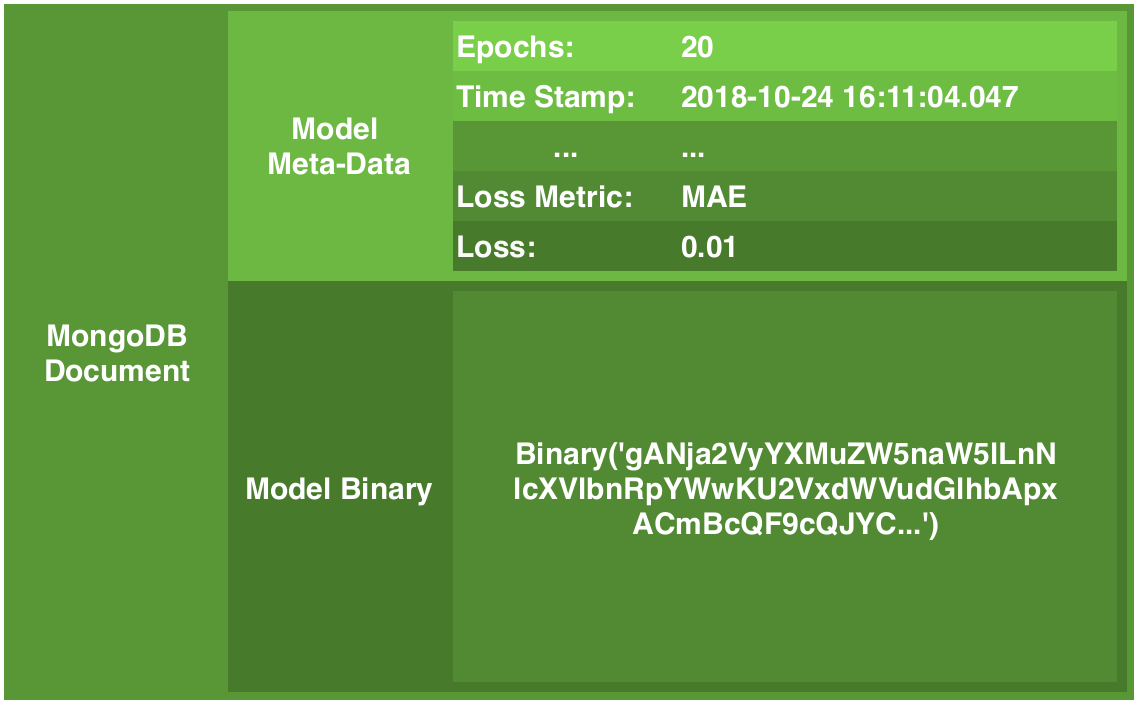}
	\caption{(colour online) MongoDB document consecutively split into abstraction hierarchy.}
    \label{fig:mongoDoc}
\end{figure}
\subsection{Time-Series Analysis}
Given the sequential nature of the signals we are dealing with in this problem, it was intuitive to utilise Recurrent Neural Networks (RNN), such as Long-Short-Term Memory (LSTM). Their cells can formulate a non linear output $a^{[t]}$ based on both the input data $x^{[t]}$ at the current time step $t$, and the previous time-step activation $a^{[t-1]}$, as described in (1), where $h$ (i.e. hyperbolic tangent) is a non-linear activation function. 
\begin{equation}
\label{eq:RNN}
a^{t} = h(x^{[t]},a^{[t-1]})
\end{equation}
 LSTM is a very popular and successful RNN architecture composed of a memory cell, an input gate, an output gate and a forget gate. The cell is able to hold in memory values over arbitrary time intervals, whereas the three gates regulate the flow of information into and out of the cell. More specifically a cell receives an input and stores it for a period of time. Intuitively this is equivalent to applying the identity function ($f(x) = x $) to the input. It is evident that the derivative of the identity function is constant; therefore when an LSTM model is trained with backpropagation through time the gradient does not vanish. Backpropagation through time is a gradient-based technique employed to train certain types of RNN models,such as LSTMs \cite{hochreiter1997long}. \par
A major problem that LSTMs circumvented over the traditional recurrent neural networks (RNN) has been exploding and vanishing gradients, which one encounters when training using hundreds of time steps. LSTMs are well suited to problems where preserving the memory of an event that occurred many time steps in the past is paramount for identifying patterns and hence informing better predictions. 
%In contrast to RNNs where error gradients vanish exponentially with the size of the time lag between important events \cite{hochreiter2001gradient}, in LSTM unit error values are back-propagated from the output, the error remains in the unit's memory. This situation means that the error is continuously fed back to each of the gates until they learn to cut off the value. 
As outlined previously, the purpose of the present study is not demonstrating or arguing the superiority of deep LSTM models, but we deploy the approach to evidence the impact that a state-of-the-art intelligent system can have in retail refrigeration systems. This study is the first step towards developing a system that can work in the background for thousands of refrigerator packs and provide candidates for DSR, whilst adhering to the requirements of the National Grid.\par 
\begin{figure}[hbt!]
    \centering
    \includegraphics[width=0.7\columnwidth]{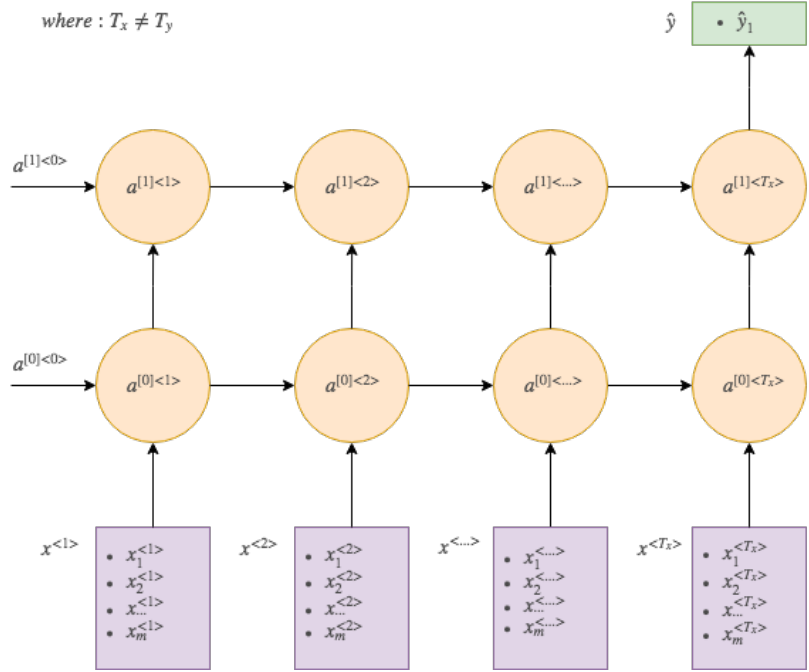}
	\caption{(colour online) Recurrent neural network architecture where: $x=$ single input example of a defrost, $x^{<n>} = $ the $n^{th}$ time step of this single example, $\hat{y} =$ is the output predictions (in this case only one per example), $a^{[n]<m>} =$ activation vector at layer n, and node/ time step m, $T_x =$ total number of time steps $T$ for examples $x$}
    \label{fig:rnn_arch}
\end{figure}
As descirbed an LSTM model was adopted due to its capability of learning long term dependencies on data. The equations relative to LSTM networks follow, and the reader is invited to refer to the original paper~\cite{hochreiter1997long} for further details.

\begin{equation}
\label{eq:LSTM} 
\begin{array}{l}
\widetilde{c}^{[t]}= tanh(W_{\tilde{c}}[a^{[t-1]},x^{[t]}]+b_{\tilde{c}})\\
\Gamma_{u}=\sigma(W_{u}[a^{[t-1]},x^{[t]}]+b_{u})\\
\Gamma_{f}=\sigma(W_{f}[a^{[t-1]},x^{[t]}]+b_{f})\\
\Gamma_{o}=\sigma(W_{o}[a^{[t-1]},x^{[t]}]+b_{o})\\
c^{[t]}=\Gamma_{u} \odot \widetilde{c}^{[t]} + \Gamma_{f} \odot c^{[t-1]} \\
a^{[t]} = \Gamma_{o} \odot tanh(c^{[t]})
\end{array}
\end{equation}

In~\ref{eq:LSTM}, $c$ is the memory cell, $\Gamma_{u}$, $\Gamma_{f}$ and $\Gamma_{o}$ are the update, forget and output gates respectively; $W$ denotes the model's weights, and $b$ are bias vectors. These parameters are all jointly learned through backpropagation. Essentially, at each time-step, a candidate update of the memory cell is proposed (\textit{i.e.} $\widetilde{c}^{[t]}$), and based on the gates $\Gamma$, $\widetilde{c}^{[t]}$ can be utilised to update the memory cell ($c^{[t]}$), and subsequently provide a non-linear activation of the LSTM cell ($a^{[t]}$). To learn a meaningful representation of the signals, two LSTMs were stacked, both with $400$ neurons (figure 4).

\subsection{Generative Adversarial Network}

Generative Adversarial Networks (GANs) proposed by Goodfellow \cite{goodfellow2014generative} are:
\begin{itemize}
    \item adversarial models (competing/ co-operating neural networks)
    \item both trained simultaneously
    \item a generator (G) neural network which seeks to simulate the data distribution as closely as possible
    \item a discriminator (D) neural network which seeks to estimate the probability that any given sample came from the training data rather than simulated by G
\end{itemize}
\begin{figure}[!hbt]
    \centering
    \includegraphics[width=0.8\columnwidth]{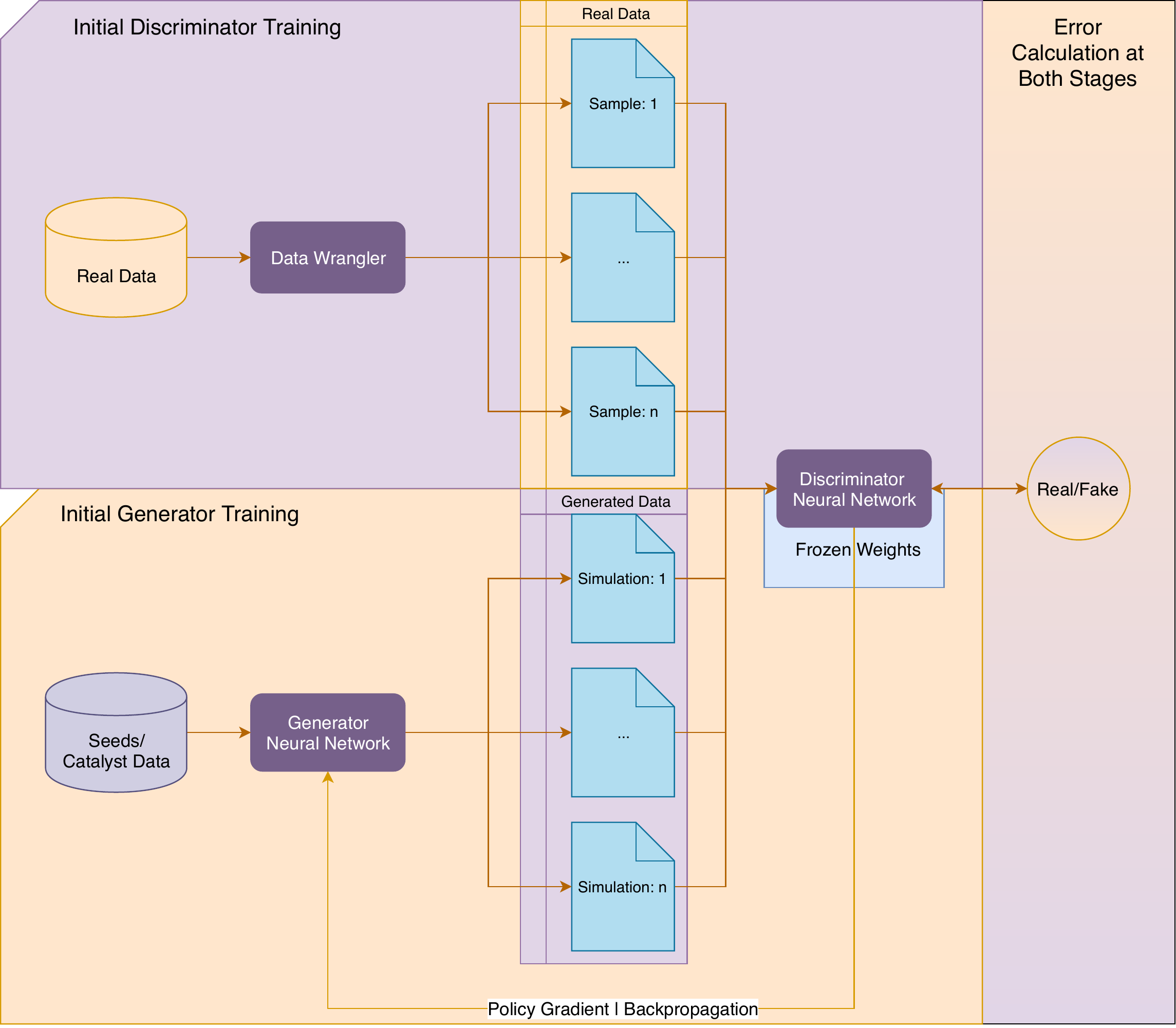}
	\caption{(colour online) General architecture/ flow of real and simulated data through a GAN.}
    \label{fig:gan}
\end{figure}
GANs (figure 5) were conceptualised through game theory of a minmax two player game, where the generator and discriminator are trying to achieve what is called the Nash Equilibrium with respect to each other. In brief, the two components of GANs, i.e. generator and discriminator, are improving themselves by competing with each other. Generator starts with random noise as input; as the process evolves it arrives to a point where it has learned the underlying distribution of the input data, hence generating a very similar output. This process continues up until the generated data are indiscernible to the original ones.
There have been advancements in the techniques used to train GANs, such as policy gradients, and the ability to predict discreet features for things, such as natural language processing or other sequence based tasks \cite{seqGan}. Using the variant of GANs called sequence-GAN or seqGAN, one could train these generative models on this predominantly sequence based/ time series prediction task.

The principles of GANs are particularly relevant to our studies, given that they are  simultaneously a type of generative and discriminative modelling compared to the soley discriminative nature of RNNs. As such GANs are very good at simulation, and provide the opportunity to expand further into data correction, since the GAN model of learning seeks to mimic the data and its distribution, which was a large hindrance using this dataset, having missing values, features (between stores). Thus well performing GANs provide us in this case study superior accuracy, and ability to replace missing data realistically. On the other hand, their primary drawbacks are that they take a more time to develop and are not at least in their vanilla form compatible with sequence data. \cite{seqGan}\\

 \section{Nemesyst for Intelligent refrigeration Systems}
 \subsection{Case Study: Demand Side Response Recommendation}
   % \subsubsection{Brief\label{subsubsec:dsr_brief}}
        Demand side response (DSR) events simply modify the power demand load in proportion to available energym on the Grid \cite{saleh2018aggregated, saleh2018impact}. However,  delivering DSR for supermarket refrigeration systems is complex, requiring high speed decisions on:

        \begin{itemize}
            \item Where to displace load with an aggregated number of machines. This requires accurate predictions/ recommendations across large pools of diverse machines
            \item Provisioning of data to train the algorithms for individual machines, in this instance the algorithm must learn the thermal inertia indirectly within each case to predict whether they can be safely shut down without a subsequent breech of food safety temperature limits during a DSR event; refrigerators with high thermal inertia can be shut down for a longer period than those with low inertia.
            \item Dealing with a large array of distributed assets, requiring a distributable algorithm or at least an algorithm that can handle database servers, which would likely be the only non-local place this data could be processed at point of need within tight time margins
            \item Time bound response, which requires both sufficient computational power and efficient algorithms to expedite the computation of this data to output within the required time-period
        \end{itemize}

        \subsubsection{Data}
        Two sources of data were used to demonstrate Nemesyst, one from the experimental refrigeration research centre at at the University of Lincoln (UoL; see for details \cite{saleh2018aggregated, saleh2018impact} and a second selected from a massive operational network (110,000 cases) of a national food retail company \cite{saleh2018aggregated, saleh2018impact}. \ref{tab:features} shows the commercial stores' refrigerator operational data features, whereas \ref{tab:data_sample} shows a sample of the data gathered at UoL, which represents a single fridge and all associated sensors in the controlled test store. 
        
\begin{figure}[ht]
    \centering
    \includegraphics[width=\columnwidth]{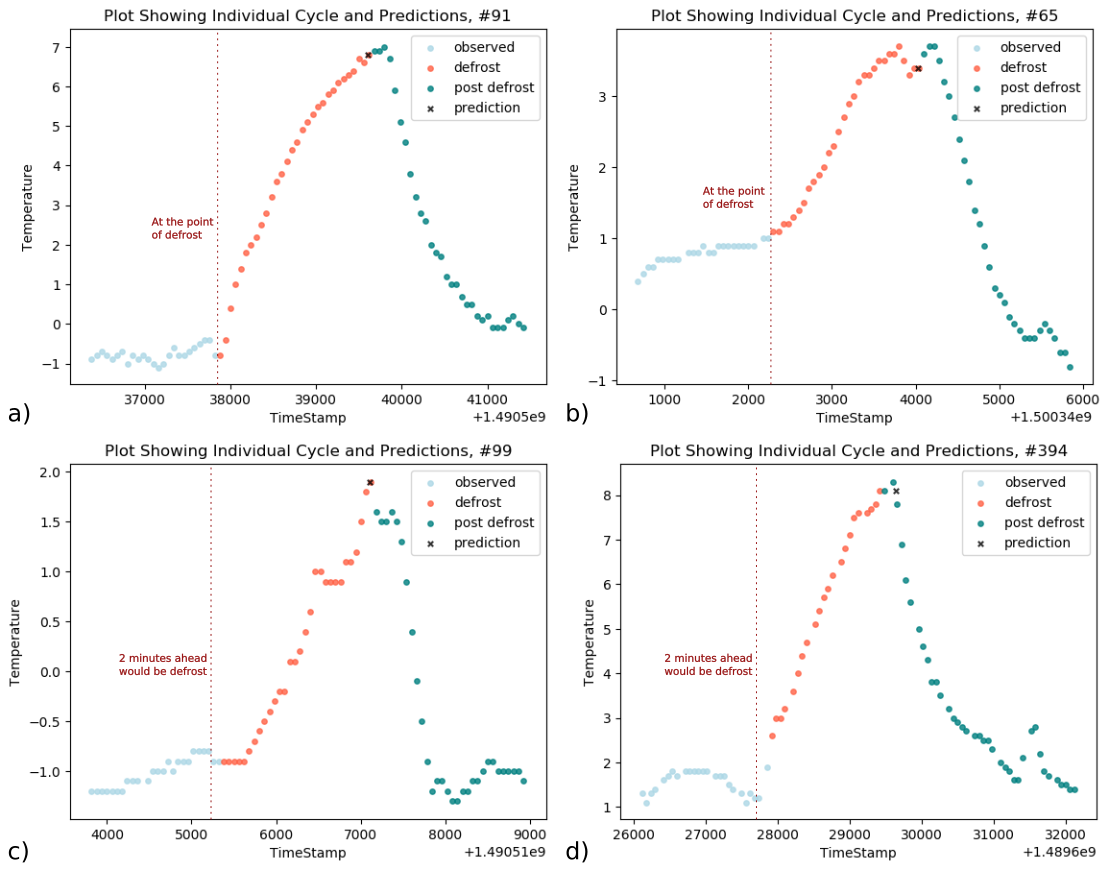}
	\caption{(colour online) Graph of air\_on\_temperature, against time, with overlayed prediction and colour coded sections for four fridges. Timestamp is simply the unix time since epoch. "Observed" (light blue) here means what the neural network is provided to learn from, "defrost" (orange) is the ground truth; of which "prediction" (x) is trying to predict the final ground truth (orange) value. The difference between the last "observed" (light blue), and final "defrost" (orange) value is the time (in seconds) that defrost can occur for before reaching the unsafe threshold of ~ 8\textdegree c. Images a) and b) show the case where the prediction takes place at the point of defrost; whereas images c) and d) show the case where the prediction occurs 2 minutes in advance of defrost: taking place.}
    \label{fig:int5}
\end{figure}
        
        Our objective was to predict the time (in seconds) until the refrigerator temperature  ("air\_on\_temperature") rises from the point it is switched off until it breaches a food safety threshold. This duration varies between cases, for example the threshold is heuristically 8\textdegree c for high temperature refrigerators, and -183\textdegree c in low temperature refrigerators (freezers), however since we do not have all the information about the threshold for each fridge or even a good way of identifying the fridge types we train the neural network on the final point of defrost (see Fig \ref{fig:int5}) from the data as this is the most consistent point we can rely on. Subsequently the variation means a subset of fridges can be selected from the population which are capable of remaining cool through a DSR event. Thus knowing how long the refrigerators can remain off before reaching this threshold allows us to identify the best candidates which reduce the power consumption sufficiently enough, while also minimising the total number of affected refrigerators, and potentially affected stock. This can be visualised in figure \ref{fig:int5} where the light blue observed data set is used to predict the duration of the defrost section (orange) before it reaches the unsafe threshold of ~8 \textdegree c (the difference between last light blue to the last orange) giving us a single predicted duration per refrigerator at any given point in time. If we used longer time periods we could potentially include information from previous defrosts but we found this unnecessary as the neural network was already more than capable with the mere fluctuations prior to defrost for prediction/ \par

        In the data from the large chain food retailer, the only indicator available on a consistent basis for how each fridge would act if it was switched off for DSR is its temperature response during a standard defrost cycle. These are routine and programmed 4 times a day for any given fridge. During a defrost cycle -- although depending on the refrigerator -- the fridge is turned off until it reaches a given temperature threshold. Case temperature data during the defrost cycles were used to train the neural networks as this was the most readily available information on refrigerator behaviour when unpowered.

        In the commercial data set, we used 110,000 defrost examples from 935 refrigerators (90,000 training, 10,000 validation, 10,000 testing), shared between 21 superstores after extraction.  When optimised the data deployed by \pfName\ was 5.1KB per data set, which results in a total data scale of ~550MB. We used only 3 months of data (limitation imposed by the data management process), and only for a subset of the 3000+ available stores i.e 21, while also eliminating redundancy and selecting certain portions of the available sequences. This will be significantly larger in practice, but is already too large to be visualised and traverse easily, another advantage of breaking this data down into usable atomic examples. The extracted valid sequences are on average 31.48 time-steps of c.1 minute long (not including missing/ empty time steps). This means we have roughly 3,462,800 time steps, each with a minimum of 7 valid data points per time step, resulting in 24,239,600 total data points processed for and subsequently used by a subset for any given neural network pre epoch. Although it should be noted an added difficulty in this scenario is that the overwhelming majority of the data is not sufficient in length or is problematic leaving on a subset that needs to be extracted for use in training. Lastly if we imagine this system will eventually be used at all stores, a simple system isolated to one machine in one store will not be sufficient to serve all stores and infer their data simultaneously. This shows the clear rationale for deploying distributing deep learning to internet of things systems, for both training and inference.
        
        % https://www.statista.com/statistics/238667/tesco-plc-number-of-outlets-worldwide/

        Data quality for the commercial stores data set posed significant problems, some of the these included:
        \begin{itemize}
            \item Missing data (features and observations). Missing features are automatically removed by Nemesyst, as they are easy to spot if the number of unique values of x are $<= 1$. Missing observations are dealt with in a customisable cleaner file in the wrangling stage as it may be desirable to maintain these observations in some form.
            \item Inconsistent data values, such as Yes, yes, y, YES etc, due to humans' inconsistent input. These are automatically dealt with at the unifying stage of the customisable cleaner file.
            \item Inconsistent number of sensors between stores and fridges, some having more and others having other inputs, making it more difficult to create a subset that share the same features/characteristics. This is dealt with in the customisable cleaner, as what is useful cannot be known and hence being able to deal with this automatically for all data.
            \item Duplicate entries, which is an expensive, time consuming operation to fix, but dealt with in the customisable cleaner.
            \item Extreme values, such as refrigerators at \SI{50.0}{\celsius} and below \SI{-40.0}{\celsius}, can be dealt with by sigma clipping in the customisable cleaner.
        \end{itemize}
        \pfName\ requires data to be separated into single/ individual examples, so that they can be imported into the MongoDB database. It is then possible to use database functionality to both parallelise  data handling, and unify the MPI of the system. As a side note, MongoDB has many tools built around it, in particular helpful visualisation tools that help speed up data exploration (figure A.1), and testing of aggregation pipelines (figure A.2).
        
    \subsubsection{Nemesyst Configuration and Methodology}
       In our public github repository,  we show the implementation and source code of the RNNs, LSTMs, and GANs implemented in this study \cite{repo_lstm}. We also show the associated RNN architecture shown in figure 3, which can show the configurability of \pfName\ through simple adaptations in arguments to shift the entire architecture. It has been further extended to other techniques, such as GANs, which are included in our implementations. It can be extended to any other novel techniques due to the separation of the learner and \pfName\ core. 
       
       In our test case, we split the 110,000 defrost examples such that  c.10,000 were used for final testing c. 10,000 were randomly selected each time for validation (Monte Carlo with replacement). We used a two-layer RNN and LSTM, using MAE (\ref{eq:mae}) as our error function and adam \cite{2014arXiv1412.6980K, j.2018on} as our optimiser. We chose MAE for this regression problem as the errors significance is equal, meaning we prefer to optimise the average case rather than penalise extreme cases.  We chose the adam optimiser for our neural networks gradient descent as it is a lightweight optimiser with a lower training cost than, AdaDelta, RMSProp, etc. The GAN contains a generator consisting of 3 layers of RNN, leaky ReLU pairs \cite{krizhevsky2012imagenet}, batch normalisation \cite{2015arXiv150203167I}, and 0.8 nesterov momentum \cite{sutskever2013importance, dozat2016incorporating}. Batch normalisation is simply the normalisation of parameter updates, between a group of examples, such as a group of size 32 will only result in a single parameter update that is some function of their constituent errors. The sequence length we settled for led to the greatest predictive power at 120 time-steps long. For the purposes of speed, we utilised our hybrid parallelisation using 3 different machines to train different models simultaneously, producing models in less than 2 minutes of training.
        
        \begin{figure}
            \centering
            \begin{equation}
                MAE = \frac{\sum\limits_{i=0}^{n-1} |y_i - \hat{y_i}|}{n}
                            \label{eq:mae}
            \end{equation}
            Mean Absolute Error (MAE), where:
            \begin{itemize}
                \item   \begin{center} 
                    $y_i$ is the $i$'th target value, and $\hat{y_i}$ is the corresponding predicted value
                        \end{center}
                \item   \begin{center}
                    $|y_i - \hat{y_i}| = |e_i|$, is the absolute difference between the actual $y_i$ and predicted $\hat{y_i}$, otherwise known as $|e_i|$
                        \end{center}
                \item   \begin{center}
                    $n$ is the total number of ground truth target values, and the total number of predicted values.
                        \end{center}
            \end{itemize}
        \end{figure}
        
        % \begin{figure}
        %     \centering
        %     \begin{equation}
        %         MSE = \frac{\sum\limits_{i=0}^{n-1} |y_i - \hat{y_i}|^2}{n}
        %                     \label{eq:mse}
        %     \end{equation}
        %     Mean Squared Error (MSE), where:
        %     \begin{itemize}
        %         \item   \begin{center} 
        %             $y_i$ is the $i$'th target value, and $\hat{y_i}$ is the corresponding predicted value
        %                 \end{center}
        %         \item   \begin{center}
        %             $|y_i - \hat{y_i}| = |e_i|$, is the absolute difference between the actual $y_i$ and predicted $\hat{y_i}$, otherwise known as $|e_i|$
        %                 \end{center}
        %         \item   \begin{center}
        %             $n$ is the total number of ground truth target values, and the total number of predicted values.
        %                 \end{center}
        %     \end{itemize}
        % \end{figure}

    \subsubsection{Results and Discussion}
\begin{table}[ht]
\scriptsize
\begin{tabular}{ccccccc}
Model Type  & Data Source  & \begin{tabular}[c]{@{}c@{}}Loss Train \\ MAE (Epochs)\end{tabular} & \begin{tabular}[c]{@{}c@{}}Loss Test \\ MAE\end{tabular} & \begin{tabular}[c]{@{}c@{}}Loss Ahead \\ 2min (Test Set)\end{tabular} & \begin{tabular}[c]{@{}c@{}}Training \\ Data Points\end{tabular} & \begin{tabular}[c]{@{}c@{}}Test \\ Data Points\end{tabular} \\  \hline
Vanilla RNN & Test Site    & 64 (20)                                                            & 85                                                       & -                                                                     & 8640                                                          & 8256                                                                                                         \\
Vanilla RNN & Single Store & 336 (23)                                                           & 640                                                      & -                                                                     & 11040                                                          & 8256                                                                                                         \\
Vanilla RNN & All Stores   & 601 (3)                                                            & 1418                                                     & -                                                                     & 84360                                                         & 11520                                                                                                       \\ \hline
LSTM        & Test Site    & \textless 10 (13)                                                  & \textless 10                                             & 23                                                                    & 8640                                                          & 8256                                                                                                         \\
LSTM        & Single Store & 87 (13)                                                            & 119                                                      & 224                                                                   & 11040                                                          & 8256                                                      \\
LSTM        & All Stores   & 286 (2)                                                            & 330                                                      & 338                                                                   & 84360                                                         & 11520                                                                                                         \\ \hline
seqGAN      & Test Site    & \textless 10 (16)                                                  & \textless 10                                             & 23                                                                    & 8640                                                          & 8256                                                                                                           \\
seqGAN      & Single Store & \textless 10 (17)                                                  & 33                                                       & 33                                                                    & 11040                                                          & 8256                                                                                                          \\
seqGAN      & All Stores   & \textless 10 (13)                                                  & 59                                                      & 15                                                                    & 84360                                                         & 11520                                                     
\end{tabular}
\caption{Table showing results for Nemesyst applied to demand side response recommendation, error is seconds from now until unsafe threshold is met. This table also shows the results for future prediction, i.e. 2 minutes ahead of potential dsr event.}
\label{tab:results}
\end{table}
\begin{figure}[!hbt]
    \centering
    \includegraphics[width=0.8\columnwidth]{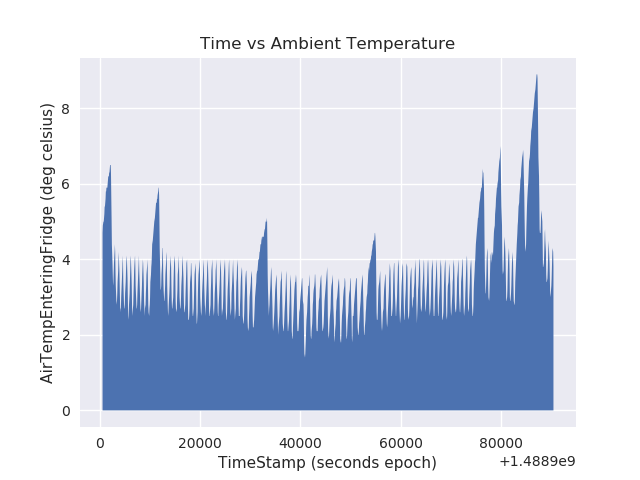}
	\caption{(colour online) Graph of air\_on\_temperature, against time for a full day.}
    \label{fig:aon}
\end{figure}

Table \ref{tab:results} shows the results achieved across the real and test stores, using all aforementioned models (bearing in mind that the data flow at a granularity of 1 minute). Single store refers to the store which returned the best performance in terms of MAE whereas All stores is the global model using data from all the stores available. It is important to note that the error is the mean absolute error (MAE) in seconds of predicted time until unsafe temperature is reached. Assuming the ground truth value of the dependent variable time (to unsafe threshold) is 1800, then we can report that one can predict this time $1800s \pm 10s$. It is worth pointing out that a normal defrosting cycle lasts between 1800 and 2700 seconds. That helps with placing the performance -- in terms of MAE -- of the presented models in Table 1 into perspective\par 
 As can be seen an expected superiority in the performance of LSTM versus vanilla RNN occur, which is down to the long   dependencies on historical data, in particular since the data of any single refrigerator has instances, denoting when they have been turned off/defrosting before (example can be seen in figure \ref{fig:aon}) . In table \ref{tab:results} we have also listed our future predictions with our developed LSTM, and GAN that show how predicting for a DSR event two minutes ahead has a minimal impact on results. In table 2,  single store performance (33 seconds in terms of MAE, using GAN) demonstrates clearly how feasible and plausible deploying such a system -- error-wise -- would be. Even using a global model for all stores the performance of 128 seconds in terms of MAE, suggests that being able to predict with such a low MAE, at 60 seconds prior to a defrosting cycle kicking in is a promising indicator. To this direction the performance we achieved in the barn store (\textless10 and 12 seconds in terms of MAE for in time and 2 min ahead predictions respectively) is a good reference point of what the absolute performance under a controlled environment can be.\par 

    Table \ref{tab:results} clearly demonstrates that training models on individual or a more tailored store specific data sets provides superior performance compared to global "all stores" data. This suggests that distributing the learning to more specific learners would be more beneficial, in particular since to train a global "all stores" model restricts the model to using only the subset of features shared between all the stores, as opposed to the more available data in one store that might not be consistent enough to use across all of them. Besides, each store exposes an underlying distinct behaviour, such as number of visitors, preference in fresh versus other products, etc., which means that a bespoke individual model for each store -- substantiated by the results provided as well -- might be the answer.\par

\begin{figure}[!t]
    \centering
    \includegraphics[width=\columnwidth]{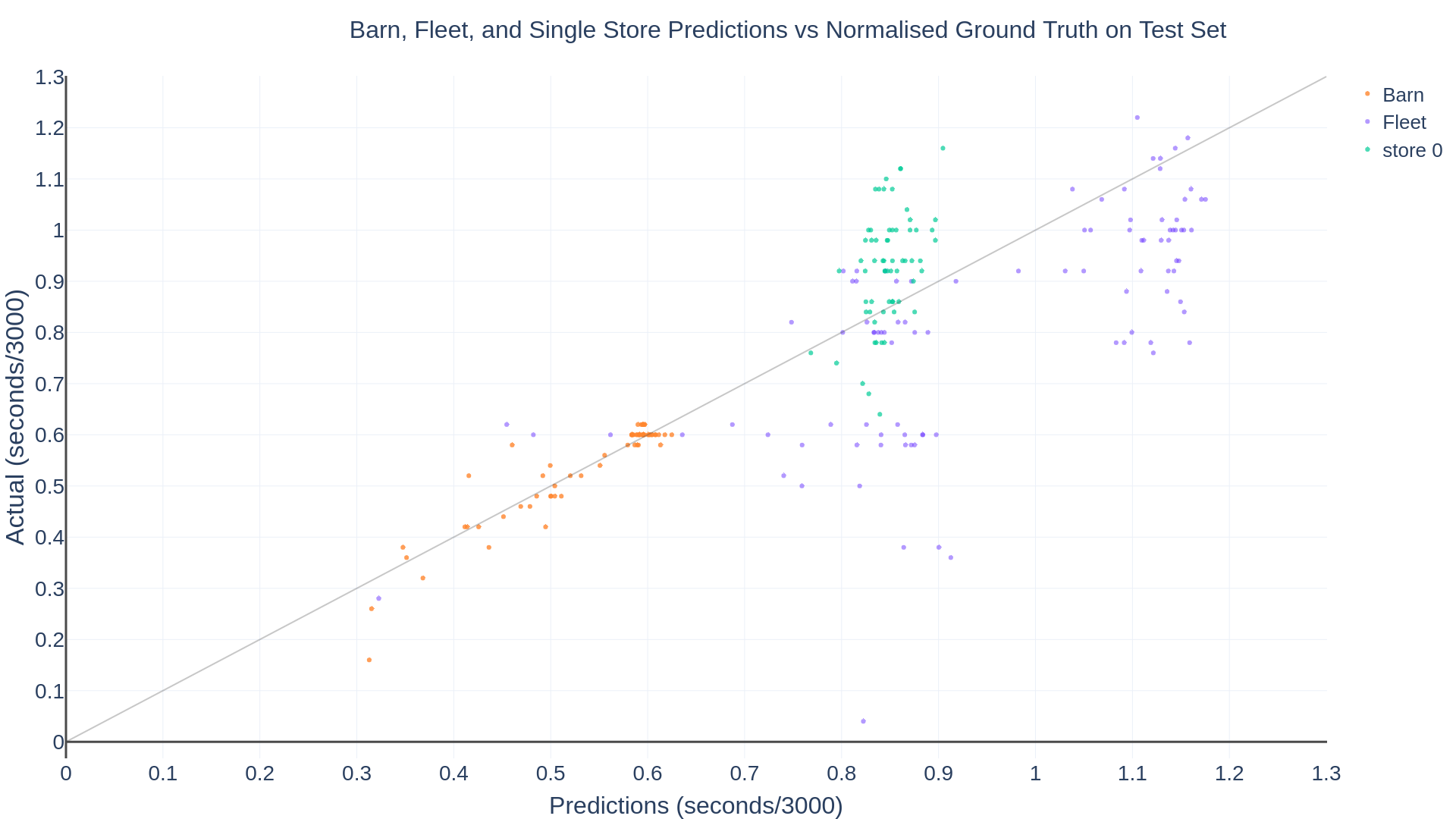}
	\caption{(colour online) Actual versus Predicted values plot, showing how close to the reference (diagonal) line the predictions are (barn, whole fleet and a random store highlighted in a different colour).  Any points falling on the diagonal line denote perfect predictions. It's worth pointing out that the predictions/dots in this plot (based on a subset of the test data) exceed the 200 ones; but many of them overlap with each other on the same spot. It is evident that the barn model performs better as can be seen in \ref{tab:results}, the reasons for this are explained in the main text. However, both models visually demonstrate how  promising the results are towards introducing an intelligent system for DSR.}
    \label{fig:predictions_truths}
\end{figure}

    The results also show a significant difference between the test site (barn) and the real sites (as evidenced by figure \ref{fig:predictions_truths} as well). Authors believe this could be due to two reasons. First, the test site has some limitations in terms of how much real-life behaviour and shopping patterns can be captured, albeit some interventions, such as opening and closing the fridges, were incorporated in the data utilised. Second and probably most importantly, the availability of complete and reliable datasets allow for a more concrete representation of the principles governing the operation of the fridges, allowing long-term patterns to be reliably identified and exploited. That is a very crucial message this paper aims at conveying, i.e. improving the availability of sensor data, investing on data science principles, such as data consistency/fidelity, establishing quality control processes, flagging of intermittent transmission of information, and generating missing values in (near) real-time, are factors that will make a massive difference in the reliability and performance of intelligent systems.\par 

    To the best of our knowledge, this is the first study to  show how an intelligent system based on a bespoke framework that can automatically handle and pre-process data, coupled with deep learning techniques can be incorporated in the workflow of IoT machine optimisation This was demonstrated for large retail supermarkets to optimise operational processes, decision-making process and subsequently reduce operational energy costs.

 \section{Conclusions and Future Work}
This paper proposed a new deep learning-based framework, i.e. \pfName\, which can provide a solution and unification of deep learning techniques using a common interface, as well as allowing hybrid parallelisation and operate within an IoT architecture. \pfName\ can store and retrieve generated models (weights, architecture), model data (train, test validation) and model metadata. As part of this work, the framework was applied to demand side response recommendation. Our proposed framework demonstrated state-of-the-art results on  predicting time until fridge temperature rises to a set point (in time and also 2 minutes ahead of defrosting cycle), which is directly applicable to demand side response recommendation in refrigeration systems. It has also been shown via comparisons between the main fleet and the barn store how slight variations in data availability and consistency can have a significant effect on the learning performance, suggesting a direct improvement in predictive performance, should data availability and fidelity be improved in the future as part of the day-to-day data aggregation process. This claim has been substantiated by several working examples using state-of-the-art models, such as LSTMs and GANs, which demonstrated that such a framework and models can be deployed to improve the scheduling of the defrosting cycles towards reacting to DSR. Lastly we have introduced \pfName, and released it openly as discussed previously so that others may use, and modify it freely, potentially contributing and furthering the project. \par

 \section*{Acknowledgements}
  This research was supported by the Innovate UK grant "The development of dynamic energy control mechanisms for food retailing refrigeration systems" with a reference number 102626. We would also like to thank our partners Intelligent Maintenance Systems Limited (IMS), the Grimsby Institute (GIFHE) and Tesco Stores Limited for their support throughout this project.
 \section*{Conflicts of Interest}
 Authors have no conflict of interest to declare.
 %\section*{Bibliography}
 \bibliographystyle{unsrt}
 \bibliography{bibliography.bib}
 \clearpage
\appendix

\counterwithin{figure}{section}
\section{Figures}
\begin{figure}[!hbt]
    \centering
    \includegraphics[width=\columnwidth]{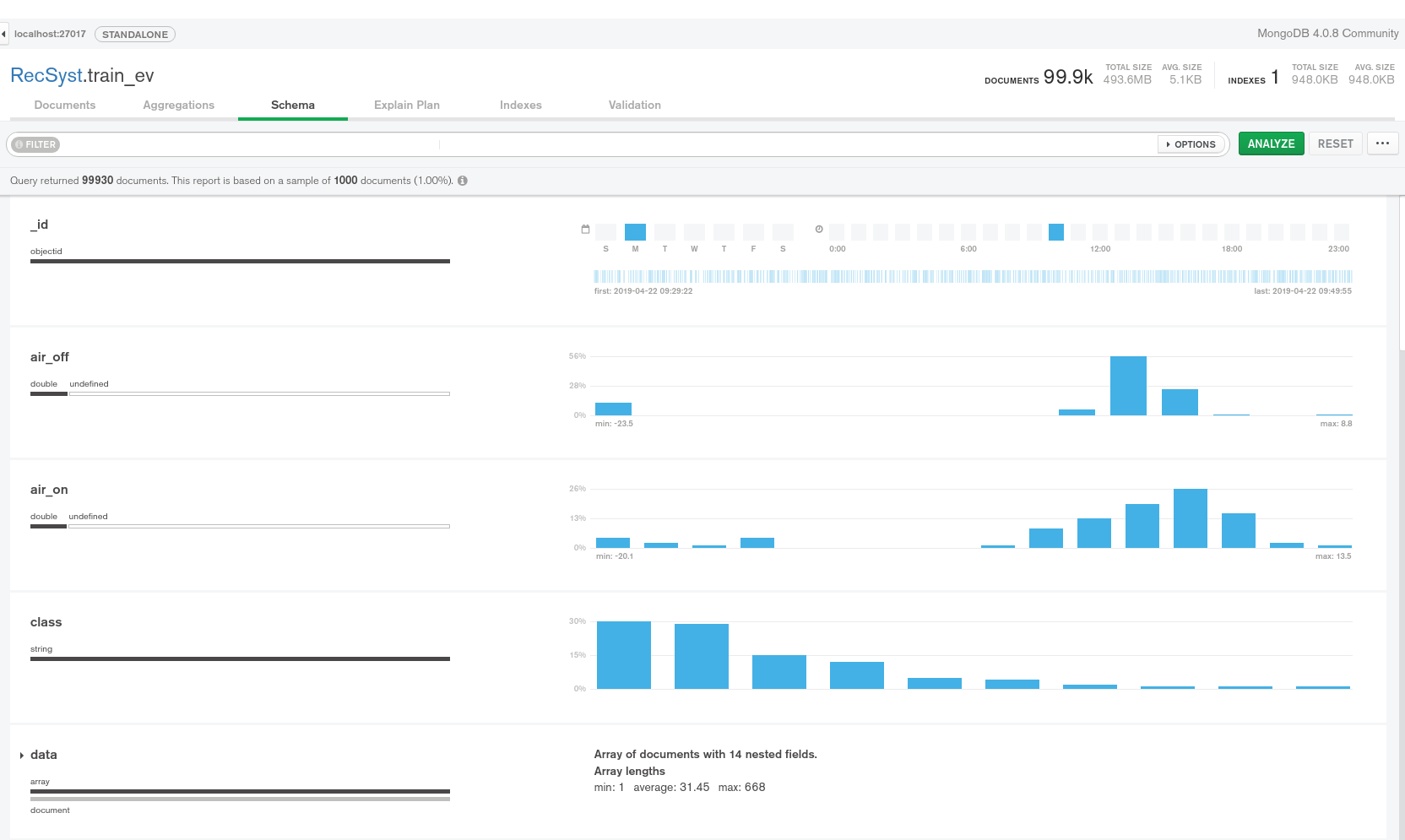}
	\caption{This figure shows MongoDB compass automatically analysing the schema for data exploration, and highlighting areas of interest, in particular here extreme temperature values that may need to either be explained or clipped.}
    \label{fig:mongo_compass1}
\end{figure}

\begin{figure}[!hbt]
    \centering
    \includegraphics[width=\columnwidth]{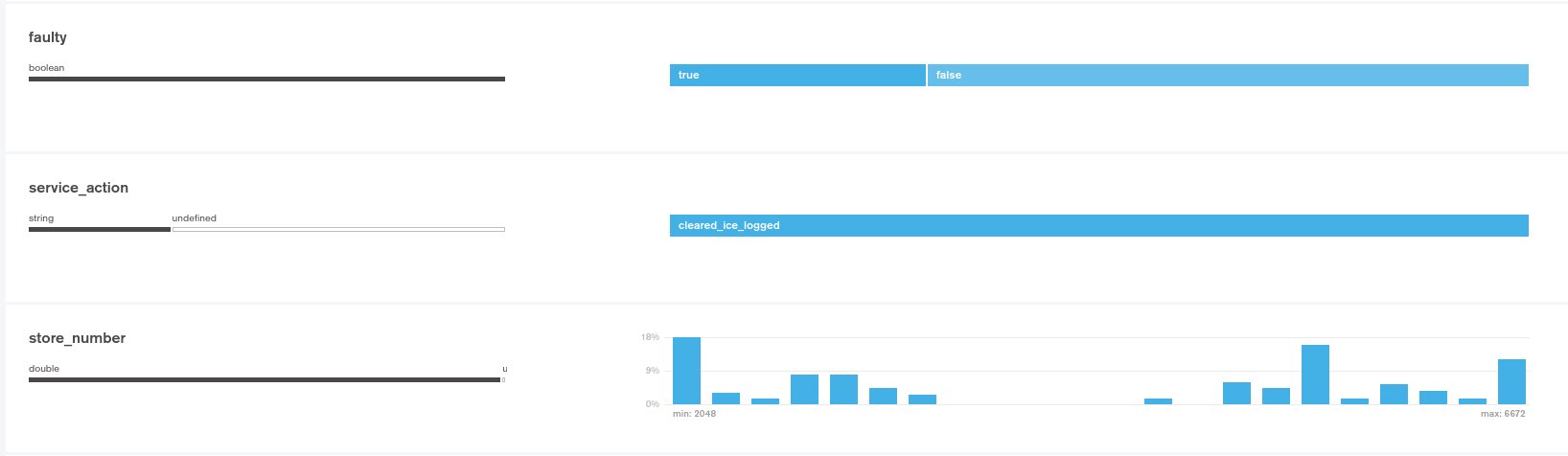}
	\caption{This figure shows MongoDB compass automatically analysing the schema for data exploration, and highlighting areas of interest, in particular an in-balanced training set, which then had to be balanced.}
    \label{fig:mongo_compass2}
\end{figure}

\begin{figure}[!hbt]
    \centering
    \includegraphics[width=\columnwidth]{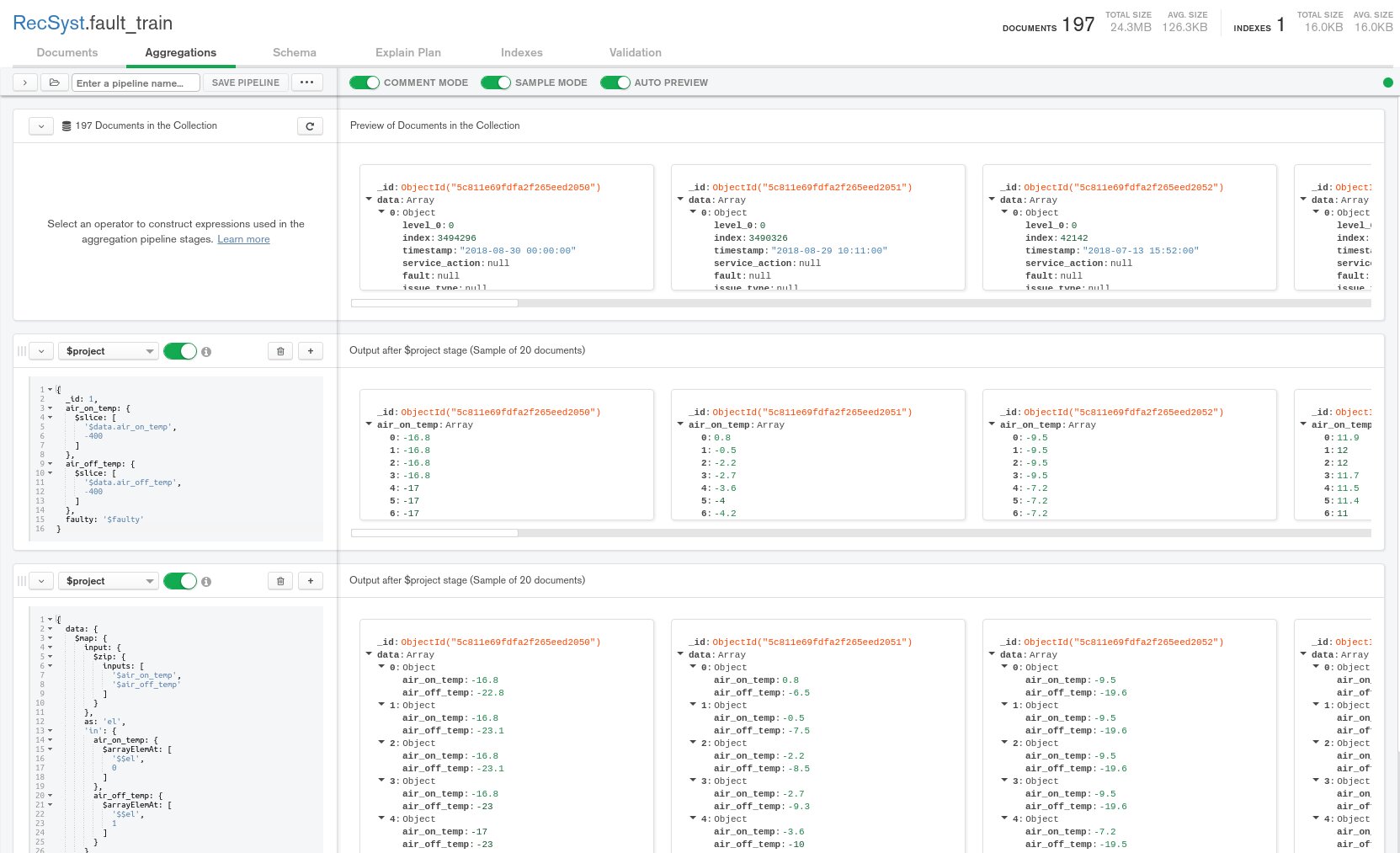}
	\caption{Example image of MongoDB aggregate pipeline testing, and output.}
    \label{fig:mongo_aggregate}
\end{figure}
\clearpage

\counterwithin{table}{section}
\section{Tables}

        \label{sec:tempData}
        \begin{table}[ht]
\centering
\begin{tabular}{rll}
\hline
Feature                                    & Barn & \multicolumn{1}{c}{Fleet} \\ \hline
\multicolumn{1}{r}{defrost\_state}        & 1    & 1                      \\
\multicolumn{1}{r}{air\_on\_temperature}  & 1    & 1                      \\
\multicolumn{1}{r}{air\_off\_temperature} & 1    & 1                      \\
\multicolumn{1}{r}{name}                  & 1    & 1                      \\
\multicolumn{1}{r}{refrigeration\_case}   & 1    & 1                      \\
\multicolumn{1}{r}{class}                 & 1    & 1                      \\
\multicolumn{1}{r}{severity}              & 1    & 1                      \\
\multicolumn{1}{r}{state}                 & 1    & 1                      \\
\multicolumn{1}{r}{ip}                    & 0    & 1                      \\
\multicolumn{1}{r}{store\_name}           & 0    & 1                      \\
\multicolumn{1}{r}{store\_number}         & 0    & 1                      \\
\multicolumn{1}{r}{timestamp}             & 1    & 1                      \\
\multicolumn{1}{r}{derived\_shelfTemp}    & 1    & 0                      \\
\multicolumn{1}{r}{derived\_foodTemp}     & 1    & 0                      \\
\multicolumn{1}{r}{evaporator\_in}        & 1    & 0                      \\
\multicolumn{1}{r}{evaporator\_out}       & 1    & 0                      \\
\multicolumn{1}{r}{evaporator\_valve}     & 1    & 0                      \\
\multicolumn{1}{r}{door\_state}           & 1    & 0                      \\ \hline
\multicolumn{1}{r}{date}                  & d    & d                      \\
\multicolumn{1}{r}{targetTemp\_on}        & d    & d                      \\
\multicolumn{1}{r}{targetTemp\_on\_diff}  & d    & d                      \\
\multicolumn{1}{r}{targetTemp\_off}       & d    & d                      \\
\multicolumn{1}{r}{targetTemp\_off\_diff} & d    & d                      \\
\multicolumn{1}{r}{targetTime}            & d    & d                      \\
\multicolumn{1}{r}{targetTime\_sec}       & d    & d                      \\
\multicolumn{1}{r}{timestamp\_sec}        & d    & d                      \\
time\_diff\_sec                            & d    & d                     
\end{tabular}
\caption{Table showing feature prevalence between data sets, where d is derived, 1 is present and 0 is absent.}
\label{tab:features}
\end{table}

\begin{landscape}
\begin{table}[h]
\begin{normalsize}
\begin{tabular}{lllllllllllllllll}
TimeStamp    & air\_on & air\_off & foodTmp & shelfTmp & Def  & Ctrl & dSP  & EvIn & EvOut & EvVlv & Def1 & Cln & Door & Lgt & Is & Is1 \\ \hline
1488990480.0 & 3.8     & 1.7      & 3.8     & 2.5      & 17.2 & 1.7  & -0.3 & -9.4 & 3.5   & 100.0 & 0    & 0   & 0    & 1   & 1  & 0   \\
1488990420.0 & 4.2     & 2.7      & 3.8     & 3.3      & 17.1 & 2.7  & 0.7  & -9.0 & 3.7   & 100.0 & 0    & 0   & 0    & 1   & 1  & 0   \\
1488990300.0 & 4.3     & 3.2      & 3.8     & 3.6      & 17.2 & 3.2  & 1.2  & -7.2 & 3.5   & 100.0 & 0    & 0   & 0    & 1   & 1  & 1   \\
1488990240.0 & 4.1     & 2.8      & 3.8     & 3.3      & 17.2 & 2.8  & 0.8  & -3.9 & 3.2   & 0.0   & 0    & 0   & 0    & 1   & 0  & 0   \\
1488990180.0 & 3.8     & 2.3      & 3.8     & 2.9      & 17.3 & 2.3  & 0.3  & -5.0 & 3.0   & 0.0   & 0    & 0   & 0    & 1   & 0  & 0   \\
1488990120.0 & 3.5     & 1.8      & 3.9     & 2.4      & 17.3 & 1.8  & -0.2 & -4.8 & 2.8   & 0.0   & 0    & 0   & 0    & 1   & 0  & 0   \\
1488990060.0 & 3.3     & 1.1      & 3.9     & 1.9      & 17.3 & 1.1  & -0.9 & -6.0 & 2.7   & 0.0   & 0    & 0   & 0    & 1   & 0  & 0   \\
1488989940.0 & 3.1     & 0.5      & 3.9     & 1.5      & 17.4 & 0.5  & -1.5 & -8.1 & 2.6   & 0.0   & 0    & 0   & 0    & 1   & 0  & 0   \\
1488989880.0 & 3.0     & 0.1      & 3.9     & 1.2      & 17.5 & 0.1  & -1.9 & -7.3 & 2.6   & 0.0   & 0    & 0   & 0    & 1   & 0  & 0   \\
1488989820.0 & 3.1     & -0.1     & 4.0     & 1.1      & 17.5 & -0.1 & -2.1 & -8.3 & 2.7   & 0.0   & 0    & 0   & 0    & 1   & 0  & 0   \\
1488989760.0 & 3.5     & 0.8      & 4.0     & 1.8      & 17.4 & 0.8  & -1.2 & -8.6 & 3.2   & 0.0   & 0    & 0   & 0    & 1   & 0  & 0   \\
1488989700.0 & 3.9     & 1.6      & 4.0     & 2.5      & 17.4 & 1.6  & -0.4 & -8.5 & 3.5   & 100.0 & 0    & 0   & 0    & 1   & 1  & 0   \\
1488989580.0 & 4.3     & 2.4      & 4.0     & 3.1      & 17.3 & 2.4  & 0.4  & -9.6 & 3.8   & 100.0 & 0    & 0   & 0    & 1   & 1  & 0   \\
1488989520.0 & 4.5     & 3.1      & 4.1     & 3.6      & 17.4 & 3.1  & 1.1  & -7.6 & 3.7   & 100.0 & 0    & 0   & 0    & 1   & 1  & 1   \\
1488989460.0 & 4.3     & 2.9      & 4.1     & 3.4      & 17.2 & 2.9  & 0.9  & -4.3 & 3.4   & 0.0   & 0    & 0   & 0    & 1   & 0  & 0   \\
1488989400.0 & 4.0     & 2.4      & 4.1     & 3.0      & 17.1 & 2.4  & 0.4  & -4.9 & 3.3   & 0.0   & 0    & 0   & 0    & 1   & 0  & 0   \\
1488989340.0 & 3.7     & 1.6      & 4.1     & 2.4      & 16.8 & 1.6  & -0.4 & -5.7 & 3.0   & 0.0   & 0    & 0   & 0    & 1   & 0  & 0   \\
% 1488989280.0 & 3.5     & 1.1      & 4.1     & 2.0      & 16.5 & 1.1  & -0.9 & -6.9 & 3.0   & 0.0   & 0    & 0   & 0    & 1   & 0  & 0   \\
% 1488989160.0 & 3.4     & 0.5      & 4.1     & 1.6      & 16.5 & 0.5  & -1.5 & -6.6 & 2.9   & 0.0   & 0    & 0   & 0    & 1   & 0  & 0  
\end{tabular}
\caption{Sample of partially cleaned data from refrigerator operation (BARN)}
\label{tab:data_sample}
\end{normalsize}
\end{table}
\end{landscape}

\begin{table}[ht]
        \centering
        \begin{tabular}{rll}
        \hline
        Feature                                    & Barn & \multicolumn{1}{c}{Fleet} \\ \hline
        \multicolumn{1}{r}{refrigeration\_case}   & 0    & 1                      \\
        \multicolumn{1}{r}{store\_number}         & 0    & 1                      \\
        \multicolumn{1}{r}{timestamp}             & 0    & 1                      \\
        \multicolumn{1}{r}{fault\_name}                 & 0    & 1                      \\
        \end{tabular}
        \caption{Table showing feature prevalence between data sets for work order faults, where d is derived, 1 is present and 0 is absent.}
        \label{tab:features_fault}
        \end{table}

%\begin{figure}[!hbt]
 %   \centering
  %  \includegraphics[width=\columnwidth]{time_priorCalShelfTemp}
%	\caption{Graph of calculated shelf temperature, against time for a full day.}
 %   \label{fig:cps}
%\end{figure}

%\begin{figure}[!hbt]
    %\centering
    %\includegraphics[width=\columnwidth]{time_CPT}
	%\caption{Graph of calculated product temperature, against time for a full %day.}
 %   \label{fig:cpt}
%\end{figure}
\end{document}